\documentclass[journal]{IEEEtran}
\ifCLASSINFOpdf
\else
   \usepackage[dvips]{graphicx}
\fi
\usepackage{url}
\usepackage{graphicx}
\usepackage{times}
\usepackage{epsfig}
\usepackage{amsmath}
\usepackage{amssymb}
\usepackage{xcolor}
\ifCLASSINFOpdf
\else
\fi
\begin{document}
%
\title{DotFAN: A Domain-transferred Face Augmentation Network for Pose and Illumination Invariant Face Recognition}
%
%
%

\author{Hao-Chiang~Shao,~\IEEEmembership{Member,~IEEE,}
        Kang-Yu~Liu,
        Chia-Wen~Lin$^\dagger$,~\IEEEmembership{Fellow,~IEEE,}
        and~Jiwen~Lu~\IEEEmembership{Senior Member,~IEEE}
\thanks{Hao-Chiang Shao is with the Department of Statistics and Information Science, Fu Jen Catholic University, Taiwan. (e-mail:shao.haochiang@gmail.com)}
\thanks{Kang-Yu~Liu is with the Department of Electrical Engineering, National Tsing Hua University, Hsinchu, Taiwan.}	
\thanks{Chia-Wen Lin (corresponding author) is with the Department of Electrical Engineering and the Institute of Communications Engineering, National Tsing Hua University, Hsinchu, Taiwan. (e-mail: cwlin@ee.nthu.edu.tw)}%
\thanks{Prof. Jiwen Lu is with Tsinghua University, Beijing, China. (e-mail: lujiwen@tsinghua.edu.cn)}
\thanks{Manuscript uploaded to arXiv on Feb. 23rd, 2020.}}

%
%

\markboth{Bare Demo of IEEEtran.cls for IEEE Journals}%
{Shell \MakeLowercase{\textit{et al.}}: Bare Demo of IEEEtran.cls for IEEE Journals}
%



\maketitle

\begin{abstract}
The performance of a convolutional neural network (CNN) based face recognition model largely relies on the richness of labelled training data. 
Collecting a training set with large variations of a face identity under different poses and illumination changes, however, is very expensive, making the diversity of within-class face images a critical issue in practice. 
In this paper, we propose a 3D model-assisted domain-transferred face augmentation network (DotFAN) that can generate a series of variants of an input face based on the knowledge 
distilled from existing rich face datasets collected from other domains. 
DotFAN is structurally a conditional CycleGAN 
but has two additional subnetworks, namely face expert network (FEM) and face shape regressor (FSR), for latent code control. While FSR aims to extract face attributes, FEM is designed to capture a face identity. With their aid, DotFAN can learn a disentangled face representation and  effectively generate face images of various facial attributes 
while preserving the identity of augmented faces.
Experiments show that DotFAN is beneficial for augmenting small face datasets to improve their within-class diversity so that a better face recognition model can be learned from the augmented dataset. 
\end{abstract}


%
\IEEEpeerreviewmaketitle

\section{Introduction}
\label{sec:intro}

%
%


\begin{figure}[!t]
\centering
\includegraphics[width=0.48\textwidth,keepaspectratio=true]{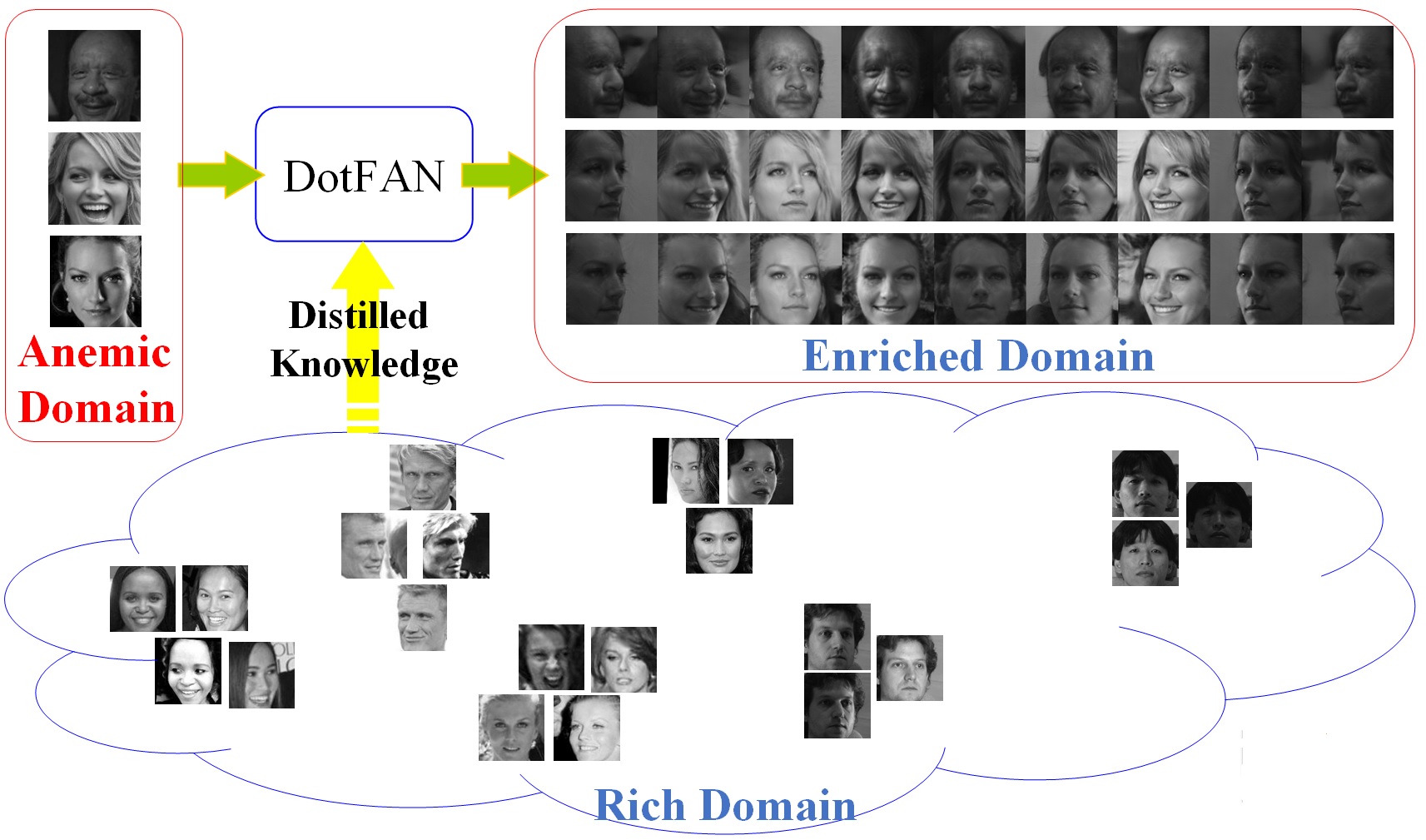} 
\caption{DotFAN aims to enrich an anemic domain via identity-preserving face generation based on the knowledge, i.e., disentangled facial representation, distilled from data in a rich domain. 
}
\label{fig:FAMteasingfigure}
\end{figure}

Face recognition is one of the most considerable research topics in the field of computer vision. Benefiting from  
meticulously-designed CNN architectures and loss functions \cite{he2016deep,deng2019arcface,wang2018cosface}, the performance of face recognition models have been significantly advanced. 
The performance of a CNN-based  face recognition model largely relies on the richness of labeled training data.
However, collecting a training set with large variations of a face identity under different poses and illumination changes is very expensive, making the diversity of within-class face images a critical issue in practice. 
A face recognition model may fail, if a test face contains what the model did not learn from an anemic training set. 
To avoid this circumstance, our idea is to distill the knowledge within a rich data domain and then transfer the distilled knowledge to enrich an incomprehensive set of training samples in a target domain  via domain-transferred augmentation. 
Specifically, we aim to train a composite network, which learns a disentangled representation of facial  attributes from rich face datasets, so that this network can generate face variants---each associating with a different pose angle, a shadow due to different illumination condition,
or a different facial expression---of each face subject in an anemic dataset for the data augmentation purpose. 
Hence, we propose a \textbf{Domain-transferred Face Augmentation Network (DotFAN)}, whose design concept is illustrated in Fig. \ref{fig:FAMteasingfigure}. 

We regard the proposed DotFAN as a 
face augmentation approach in which any identity class---no matter a minority class or not---can be enriched by synthesizing face samples based on the knowledge learned from rich face datasets in other domains via domain transfer. 
To this end,  DotFAN first learns a disentangled facial representation, through which the face information can be spanned by various face attribute codes, from rich datasets. Then, exploiting the disentangled facial representation, DotFAN can generate synthetic face samples neighboring to the input faces in the sample space so that the diversity of each face-identify class can be significantly enhanced. As a result, the performance of a face recognition model trained on the enriched dataset can be improved as well. 

%
%
%

Utilizing two auxiliary subnetworks, namely a face-expert model (FEM) \cite{cole2017synthesizing, qian2019unsupervised} and a face shape regressor (FSR), DotFAN operates intrinsically in a 3D model-assisted data-driven fashion.
FEM is a purely data-driven subnetwork pretrained on a domain rich in face identities, whereas FSR is driven by a 3D face model and pretrained on another domain with rich poses and expressions. 
Hence, FSM ensures that synthesized vaiants of an input face are of the same identity as the input, while FSR collaborating with illumination code offers the model for synthesizing faces with various poses, lighting (shadow) conditions, and expressions.\footnote{Although DotFAN can synthesize faces with various expressions, we do not particularly consider it in this work, as we do not find a suitable labeled training set with rich expressions.} In addition, inspired by FaceID-GAN~\cite{shen2018faceid}, we used the 3D face model (e.g., 3DMM \cite{blanz1999morphable}) to characterize face attributes with only hundreds of parameters. Thereby, the size of FSR, and its training set as well, is largely reduced, making it realizable with a light CNN. Furthermore, the loss terms related to FEM and FSR act as regularizers during the training stage. This design prevents DotFAN from common issues in data-driven approaches, e.g. overfitting due to small training dataset.   

Moreover, DotFAN is distinguishable from FaceID-GAN because of following reasons.
First, based on a 3-player game strategy, FaceID-GAN regards its face-expert model as an additional discriminator that needs to be trained jointly with its generator and discriminator in an adversarial training manner. Because its face-expert model assists its discriminator rather than its generator, FaceID-GAN guarantees only the upper-bound of identity-dissimilarity. This design also prevents its face expert model from pretraining and impedes the whole training speeds. Furthermore, since it cannot be pretrained on a rich-domain data, this makes it very difficult to do knowledge transfer from a rich dataset to  another  dataset in an on-line learning manner.
On the contrary, DotFAN regards its FEM as a regularizer to guarantee that the identity information is not altered by the generator. Accordingly, FEM can be pretrained on a rich dataset and play a role of an inspector in charge of overseeing identity-preservability. This design not only carries out the identity-preserving face generation task, but also stabilizes and speeds up the training process by not intervening the competition between generator and discriminator.

The main contributions of DotFAN are threefold.
\begin{itemize}
    \item We are the first to propose a domain-transferred face augmentation scheme that can easily transfer the knowledge distilled from a rich domain to an anemic domain, while preserving the identity of augmented faces in the target domain.  
    \item Through well disentangled facial representation learned from existing face data, DotFAN offers a unique unified framework that can incorporate prominent face attributes (pose, illumination, shape, expression) for face recognition and can be easily extended to other face related tasks.  
    \item DotFAN well beats the state-of-the-arts by a significant gain margin in face recognition application with  small-size training data available. This makes it a powerful tool for low-shot learning applications.  
\end{itemize}
\section{Related Work}
\label{sec:review}


Recently, various algorithms have been proposed to address the issue of small sample size with dramatic variations in facial attributes in face recognition~\cite{choi2018stargan,lu2018attribute,li2018beautygan,shen2017learning}.
This section reviews works on GAN-based image-to-image translation, face generation, and face frontalization/rotation techniques related to face augmentation.

\noindent \textbf{(A) GAN-based image-to-image translation:} 

GAN and its variants have been widely adopted in a variety of fields, including image super-resolution, image synthesis, image style transfer, and domain adaptation. 
DCGAN~\cite{radford2015unsupervised} incorporates deep CNNs into GAN for unsupervised representation learning. DCGAN enables arithmetic operations in the feature space so that face synthesis can be controlled by manipulating attribute codes. 
The concept of generating images with a given condition has been adopted in succeeding works, such as Pix2pix~\cite{isola2017image} and CycleGAN~\cite{zhu2017unpaired}. Pix2pix requires pair-wise training data to derive the translation relationship between two domains, whereas CycleGAN relaxes such limitation and exploits unpaired training inputs to achieve domain-to-domain translation.
After CycleGAN, StarGAN \cite{choi2018stargan} addresses the multi-domain image-to-image translation issue. 
With the aids of a multi-task learning setting and a design of domain classification loss, StarGAN's discriminator minimizes only the classification error associated to a known label. As a result, the domain classifier in the discriminator can guide the generator to learn the differences among multiple domains.
Recently, an attribute-guided face generation method based on a conditional CycleGAN was proposed in \cite{lu2018attribute}. This method synthesizes a high-resolution face based on an low-resolution reference face and an attribute code extracted from another high-resolution face. 
%
%
Consequently, by regarding faces of the same identity as one sub-domain of faces, we deem that face augmentation can be formulated as a multi-domain image-to-image translation problem that can be solved with the aid of attribute-guided face generation strategy. 

\begin{figure*}[!t]
\vspace{-0.15in}
\begin{tabular}{p{400pt}p{50pt}}
	\includegraphics[width=0.95\textwidth]{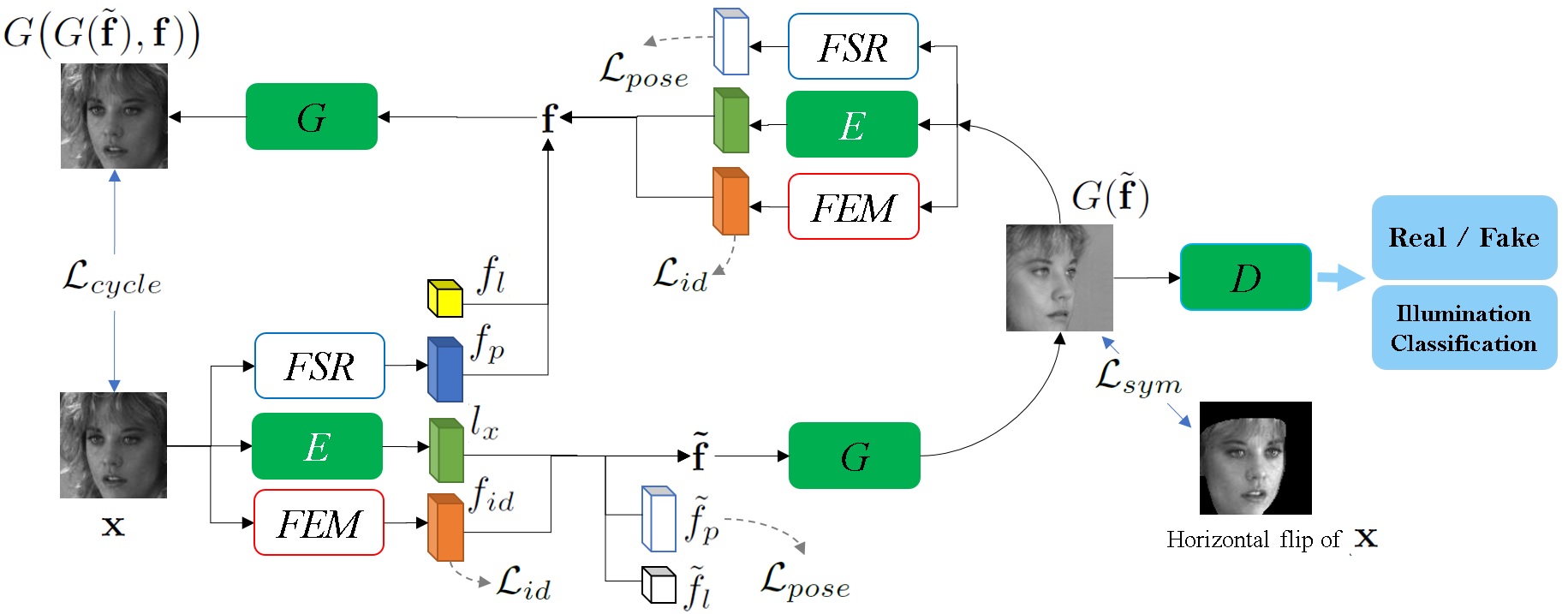}
	\par & \\[-0.5cm]
\end{tabular} 
	\caption{Training flow of DotFAN. FEM and FSR are independently pre-trained subnetworks, 
	 whereas $E$, $G$, and $D$ are trained as a whole. 
	$\tilde{f_p}$ and $\tilde{f_l}$ denote respectively a pose code and an illumination code randomly given in the training routine; and, $f_l$ is the ground-truth illumination code provided by the training set. Note that for inference, the data flow begins from $\mathbf{x}$ and ends at $G(\tilde{\mathbf{f}})$. 
	}
	\label{fig:arch}
\end{figure*}

\noindent \textbf{(B) Face frontalization and rotation:}

%

We regard the identity-preserving face rotation task as an inverse problem of the face frontalization technique used to synthesize a frontal face from a face image with arbitrary pose variation.
Typical face frontalization and rotation methods synthesize a 2D face via 3D surface model manipulation, including pose angle control and facial expression control, such as FFGAN \cite{yin2017towards} and FaceID-GAN \cite{shen2018faceid}. 
%
%
%
%
Still, some designs utilize specialized sub-networks or loss terms to reach the goal. For example, based on TPGAN \cite{huang2017beyond}, the pose invariant module (PIM) proposed in \cite{zhao2018towards} contains an identity-preserving frontalization sub-network and a face recognition sub-network; the CNN proposed in \cite{zhang2018face} establishes a dense
correspondence between paired non-frontal and frontal faces; and, the face normalization model (FNM) proposed in \cite{qian2019unsupervised} involves a face-expert network, a pixel-wise loss, and a face attention discriminators to generate a faces with canonical-view and neutral expression.
Finally, some methods approached this issue by means of disentangled representations, such as DR-GAN~\cite{tran2017disentangled} and CAPG-GAN~\cite{hu2018pose}. The former utilizes an encoder-decoder structure to learn a disentangled representation for face rotation, whereas the latter adopts a two-discriminator framework to learn simultaneously pose and identity information.


\noindent \textbf{(C) Data augmentation for face recognition:}\\
To facilitate face recognition, there are several face normalization and data augmentation methods. Face normalization methods aim to align face images by removing the volatility resulting from illumination variations, changes of facial expressions, and different pose angles \cite{qian2019unsupervised}, whereas the data augmentation method attempts to increase the richness of face images, often in aspects of pose angle and illumination conditions, for the training routine. 
To deal with illumination variations, 
conventional approaches utilized either physical models, e.g. Retinex theory \cite{land1971lightness}, or 3D reconstruction strategy to remove/correct the shadow on a 2D image \cite{finlayson2002removing,wang2008face}. 
%
%
Moreover, to mitigate the influence brought by pose angles, two categories of methods were proposed, namely pose-invariant face recognition methods and face rotation methods. While the former category focuses on learning pose-invariant features from a large-scale dataset \cite{masi2016pose,cao2018pose}, the latter category, including face frontalization techniques, aims to learn the relationship between rotation angle and resulting face image via a generative model \cite{yin2017towards,huang2017beyond,zhao2018towards,tran2017disentangled,hu2018pose,shen2018faceid}. Because face rotation methods are designed to increase the diversity of the view-points of face image data, they are also beneficial for augmentation tasks.

Based on these meticulous designs, DotFAN is implemented as a uni-generator conditional CycleGAN, involving an encoder-decoder framework and two sub-networks for learning disentangled attribute codes and triggered by several loss terms, such as cycle-consistency loss and domain classification loss, as will be elaborated later.

\section{Domain-Transferred Face Augmentation}
\label{sec:method}


The proposed DotFAN is a framework to synthesize face images of one domain based on the knowledge, i.e., disentangled facial representation, learned from others.
For a given input face $\mathbf{x}$, the generator $G$ of DotFAN is trained to synthesize a face $G(\mathbf{f})$ 
based on an input attribute code $\mathbf{f}$ comprising 
i) a general latent code $l_\mathbf{x}=E(x)$ extracted from $\mathbf{x}$ by the general facial encoder, 
ii) an identity code $f_{id}$ indicating the face identity, 
iii) an attribute code $f_p$ describing facial attributes including pose angle and facial expressions, and 
iv) an illumination code $f_l$. 
Through this design, a face image can be embedded via a disentangled representation in an attribute code $\mathbf{f}=[l_\mathbf{x}, f_{id}, f_p, f_l]$. 
%
%
Fig. \ref{fig:arch} depicts the flow-diagram of DotFAN, where each component will be elaborated in following subsections. 
%



\subsection{Disentangled Facial Representation}
\label{sec301}

To obtain a disentangled representation, the attribute code $\mathbf{f}$ used by DotFAN for generating face variants is derived collaboratively by a general facial encoder $E$, a face-expert sub-network FEM, a shape-regression sub-network FSR, and an illumination code $f_l$.
FEM and FSR are two well pre-trained sub-networks.
FEM learns to extract identity-aware features from faces (of each identity) with various head poses and facial expressions, whereas 
FSR aims to learn pose features based on a 3D model. 
The illumination code is a $14 \times 1$ one-hot vector specifying $1$ label-free case (corresponding to data from CASIA \cite{yi2014learning}) and $13$ illumination conditions (associated with selected Multi-PIE dataset \cite{gross2010multi}).


\noindent \textbf{(A) Face-Expert Model (FEM):}
FEM, denoted by $\Phi_{fem}$, enables DotFAN to extract and to transplant the face identity from an input source to synthesized face images. 
Though conventionally face identity extraction is considered as a classification problem and optimized by using a cross-entropy loss, recent  methods, e.g., CosFace \cite{wang2018cosface} and ArcFace \cite{deng2019arcface}, proposed to adopt angular information instead. ArcFace maps face features onto a unit hyper-sphere and adjust between-class distances by using a pre-defined margin value so that a more discriminative feature representation can be obtained. 
Using ArcFace, FEM ensures not merely a fast training speed for learning face identity but also the efficiency in optimizing the whole DotFAN network.
\noindent \textbf{(B) Face Shape Regressor (FSR):}
FSR, denoted as $\Phi_{fsr}$, aims to extract face attributes including face shape, pose, and expression. A fully data-driven approach requires to learn a CNN model of high complexity to completely characterize the face attributes without a prior model, which implies the need of a large variety of labeled face samples for training, thereby running a high risk of overfitting. Instead, we use a face model-assisted CNN based on the widely adopted 3D Morphable Model (3DMM \cite{blanz1999morphable}) to significantly reduce the model size (say, a light CNN), as 3DMM can fairly accurately characterize the face attributes using only hundreds of parameters. 
We follow HPEN's strategy \cite{zhu2015high} to prepare ground-truth 3DMM parameters $\Theta_\mathbf{x}$ of a given face $\mathbf{x}$ from CASIA dataset \cite{blanz1999morphable}. Then, we train FSR via Weighted Parameter Distance Cost (WPDC) \cite{zhu2016face} defined in Eq. (\ref{eq:WPDCloss_src}), with a modified importance matrix, as shown in Eq. (\ref{eq:WPDCweight}).
\begin{eqnarray}
    \mathcal{L}_{wpdc} &=& \big( \Phi_{fsr}( \mathbf{x} ) - \Theta_\mathbf{x} \big)^t \mathbf{W} \big( \Phi_{fsr}( \mathbf{x} ) - \Theta_\mathbf{x} \big) 
    \label{eq:WPDCloss_src}\\
    \mathbf{W} &=& ( w_R, w_T, w_{shape}, w_{exp}) \mbox{,}
    \label{eq:WPDCweight}
\end{eqnarray}
where $w_R$, $w_{t_{3d}}$, $w_{shape}$, and $w_{exp}$ are the distance-based weighting coefficients for $\Theta_\mathbf{x}$ (including a $9\times1$ vectorized rotation matrix $R$, a $3\times1$ translation vector $T$, a $199 \times 1$ vector $\alpha_{shape}$, and a $29 \times 1$ $\alpha_{exp}$) derived by 3DMM. 
Note that 3DMM  expresses a face as $\mbox{S} = \bar{S} + A_{shape} \alpha_{shape} + A_{exp} \alpha_{exp}$, where $\bar{S}$ is the mean face, $A_{shape}$ denotes the PCA basis spanning shape information, $A_{exp}$ is the basis for facial expressions, and $\alpha_{shape}$ and $\alpha_{exp}$ are weighting vectors. 
While training DotFAN, $\alpha_{shape}$ that represents facial shape is unchanged, and components of the translation, rotation, and expression could be replaced by arbitrary values.

\noindent \textbf{(C) General facial encoder $E$ and illumination code $f_l$:} \\
$E$ is used to capture other features, which cannot be represented by shape and identity codes, on a face. 
$f_l$ is a one-hot vector specifying the lighting condition, based on which our model synthesizes a face. 
Note that because CASIA  has no shadow labels, for $f_l$ of a face from CASIA, its former $13$ entries are set to be $0$'s and its $14^{th}$ entry $f_l^{casia}=1$; this means to skip shading and to generate a face 
with the same illumination setting and the same shadow as the input.

%

\subsection{Generator}
\label{subsec:330}

The generator $G$ 
takes an attribute code $\mathbf{f}~=~[l_\mathbf{x},~f_{id},~f_p,~f_l]$ as its input to synthesize a face $G(\mathbf{f})$. 
Described below are loss terms composing the loss function of our generator.


\noindent \textbf{ (A) Cycle-consistency loss}: \\
In our design, we adopt the cycle-consistency loss to retain face contents after performing two transformations dual to each other. That is, 
\begin{equation}
    \mathcal{L}_{cycle} = \| G\big( G(  \tilde{\mathbf{f}}), \mathbf{f})\big)- \mathbf{x} \|_2^2 / N \mbox{,}
    \label{eq:cycleloss}
\end{equation}
where $N$ is the number of pixels, $G( \tilde{\mathbf{f}})$ is a synthetic face derived according to 
an input attribute code $\tilde{\mathbf{f}}$. This loss guarantees our generator can learn the transformation relationship between any two dual attribute codes.

%
%
%

\noindent \textbf{ (B) Pose-symmetric loss}:  \\
Based on a common assumption that a human face is symmetrical, a face with an $x^\circ$ pose angle and a face with a $-x^\circ$ angle should be symmetric about the $0^\circ$ axis. Consequently, we design a pose-symmetric loss based on which DotFAN can learn to generate $\pm x^\circ$ faces from either training sample. 
This pose-symmetric loss is evaluated with the aid of a face-mask $M(\cdot)$, which is defined as a function of 3DMM parameters predicted by FSR and makes this loss term focus on the face region by filtering out the background, as described below:

\begin{equation}
    \mathcal{L}_{sym} = \| M( \hat{\mathbf{f}}^{-} ) \cdot \big( G(\hat{\mathbf{f}}^{-}) - \hat{\mathbf{x}}^{-} \big) \|^2_2 / N \mbox{.}
\end{equation}
Here, $\hat{\mathbf{f}}^-=[l_\mathbf{x}, f_{id}, \hat{f}^-_p, f_l]$, in which $\hat{f}^{-}_{p} = \Phi_{fsr}(\hat{\mathbf{x}}^{-})$, and the other three attribute codes are extracted from $\mathbf{x}$. Additionally, $\hat{\mathbf{x}}^{-}$ is the horizontally-flipped version of $\mathbf{x}$. 
In sum, this term measures the $L_2$-norm of the difference between a synthetic face and the 
horizontally-flipped version of $\mathbf{x}$ within a region-of-interest defined by a mask $M$.

\noindent \textbf{ (C) Identity-Preserving Loss}:  \\
We adopt the following identity-preserving loss to ensure that the identity code of a synthesized face 
$G( \mathbf{\tilde{f}} )$ is identical to that of input face $\mathbf{x}$. That is,
\begin{equation}
    \mathcal{L}_{id} = \| \Phi_{fem}( \mathbf{x} ) - \Phi_{fem}\big(G( \tilde{\mathbf{f}})\big) \|^2_2 / N
    \mbox{.}
    \label{eq:idloss}
\end{equation}

\noindent \textbf{ (D) Pose-consistency loss}:  \\
This term guarantees that the pose and expression feature extracted from a synthetic face is consistent with  $\tilde{f}_p$  used to generate the synthetic face. That is, 
\begin{equation}
    \mathcal{L}_{pose} = \| \tilde{f}_p - 
    \Phi_{fsr}\big(G( \tilde{\mathbf{f}})\big) \|^2_2 / N
    \mbox{.}
    \label{eq:poseloss}
\end{equation}

\subsection{Discriminator}
\label{subsec:340}
%
%

By regarding faces of the same identity as one sub-domain of faces, the task of augmenting faces of different identities becomes a multi-domain image-to-image translation problem addressed in StarGAN \cite{choi2018stargan}. Hence, we exploit an adversarial loss to make augmented faces 
photo-realistic. To this end, we use the domain classification loss to verify if 
$G( \tilde{\mathbf{f}})$ is properly classified to a target domain label $f_l$ that we used to specify the illumination condition of $G( \tilde{\mathbf{f}})$. 
In addition, in order to stabilize the training process, we adopted the loss design used in WGAN-GP \cite{gulrajani2017improved}. Consequently, these two loss terms can be expressed as follows:
\begin{eqnarray}
    \mathcal{L}_{adv}^D &=& D_{src}( G( \tilde{\mathbf{f}}) ) - D_{src}( \mathbf{x} ) \nonumber \\
    &&+ \lambda_{gp} \cdot \big( \| \nabla_{\hat{x}} D_{src}( \hat{x} ) \|_2 -1 \big)^2 \nonumber \\
    \mathcal{L}^G_{adv} &=& -D_{src}( G( \tilde{\mathbf{f}}) ) \mbox{,}
    \label{eq:WGANGPloss}
\end{eqnarray}
where $\lambda_{gp}$ is a trade-off factor for the gradient penalty,  $\hat{x}$ is uniformly sampled from the linear interpolation between $\mathbf{x}$ and synthesized 
$G( \tilde{\mathbf{f}})$, 
and $D_{src}$ reflects a distribution over sources given by the discriminator; and,
\begin{eqnarray}
   \mathcal{L}^D_{cls} &=& - \log D_{cls}( f_l^\mathbf{x} | \mathbf{x} ) \nonumber \\
    \mathcal{L}_{cls}^G &=& - \log D_{cls} ( f_l | G( \tilde{\mathbf{f}}) ) \mbox{,}
\end{eqnarray}
where $f_l^\mathbf{x}$ is the ground-truth illumination code of $\mathbf{x}$. 
In sum, the discriminator aims to produce probability distributions over both source and domain labels, i.e., $D:~\mathbf{x} \rightarrow~\{D_{src}(\mathbf{x}),~D_{cls}(\mathbf{x})\}$. Empirically,  $\lambda_{gp}=10$.

\subsection{Full objective function}
 In order to optimize the generator and alleviate the training difficulty, we pretrained FSR and FEM with corresponding labels. Therefore, while training the generator and the discriminator, no additional label is needed. The full objective functions of DotFAN can be expressed as:
\begin{eqnarray}
    \mathcal{L}_{G} &=&
        \mathcal{L}^{G}_{adv}+
        \mathcal{L}^{G}_{cls}+
        \mathcal{L}_{id}+
	    \mathcal{L}_{pose}+
        \mathcal{L}_{sym}+
        {L}_{cycle} \nonumber \\
    \mathcal{L}_{D} &=& \mathcal{L}^{D}_{adv}+ \mathcal{L}^{D}_{cls} \mbox{.}
    \label{eq:totalloss}
\end{eqnarray}
Two loss terms in $\mathcal{L}_{D}$ are equal-weighted; and, the weighting factors of terms in $\mathcal{L}_{G}$ in turn are $1$, $1$, $8$, $6$, $5$, and $5$. 



\section{Experimental Results}
\label{sec:exp}

\subsection{Dataset}
\label{subsec:410data}

DotFAN is trained jointly on \textbf{CMU Multi-PIE} \cite{gross2010multi} and \textbf{CASIA} \cite{yi2014learning}.
Multi-PIE contains more than $750,000$ images of $337$ identities, each with 20 different sorts of illumination and 15  different poses. 
We select images of pose angles ranging in between $\pm45^\circ$ 
and illumination codes from 0 to 12 to form our first training set, containing totally $84,000$ faces. From this training set, DotFAN learns the representative features for a wide range of pose angles, illumination conditions, and resulting shadows. 
Our second dataset is the whole CASIA set that contains $494,414$ images of $10,575$ identifies, each having about 50 images of different poses and expressions. Since CASIA contains a rich collection of face identities, it helps DotFAN learn features for representing identities.

To evaluate the performance of DotFAN on face synthesis, four additional datasets are used: \textbf{LFW}~\cite{huang2008labeled}, \textbf{IJB-A}~\cite{klare2015pushing}, 
\textbf{SurveilFace-1}, and \textbf{SurveilFace-2}. 
LFW  has $13,233$ images of $5,749$ identities;
IJB-A has $25,808$ images of $500$ identities;
SurveilFace-1 has $1,050$ images of $73$ identities; and SurveilFace-2 contains $1,709$ images of $78$ identities. 
We evaluate the performance of DotFAN's face frontalization on LFW and IJB-A. Besides, because faces in two SurveilFace datasets are taken in uncontrolled real working environments, they are contaminated by strong backlight, motion blurs, extreme shadow conditions, or influences from various viewpoints. Hence, they mimic the real-world conditions and thus are suitable for evaluating the face augmentation performance. 
The two SurveilFace sets are private data provided by a video surveillance provider. We will make them publicly available after removing personal labels. 

We exploit CelebA to simulate the data augmentation process. CelebA contains $202,599$ images of $10,177$ identities with 40 kinds of diverse
binary facial attributes. We randomly select a fixed number of images of each face identity from CelebA to form our simulation set, called ``\textbf{sub-CelebA}" and conducted data augmentation experiments on both CelebA and sub-CelebA by using DotFAN.

%
%
%
%
\begin{figure}[t]
\begin{tabular}{p{220pt}p{1pt}}
\centering
\includegraphics[width=0.35\textwidth,keepaspectratio=true]{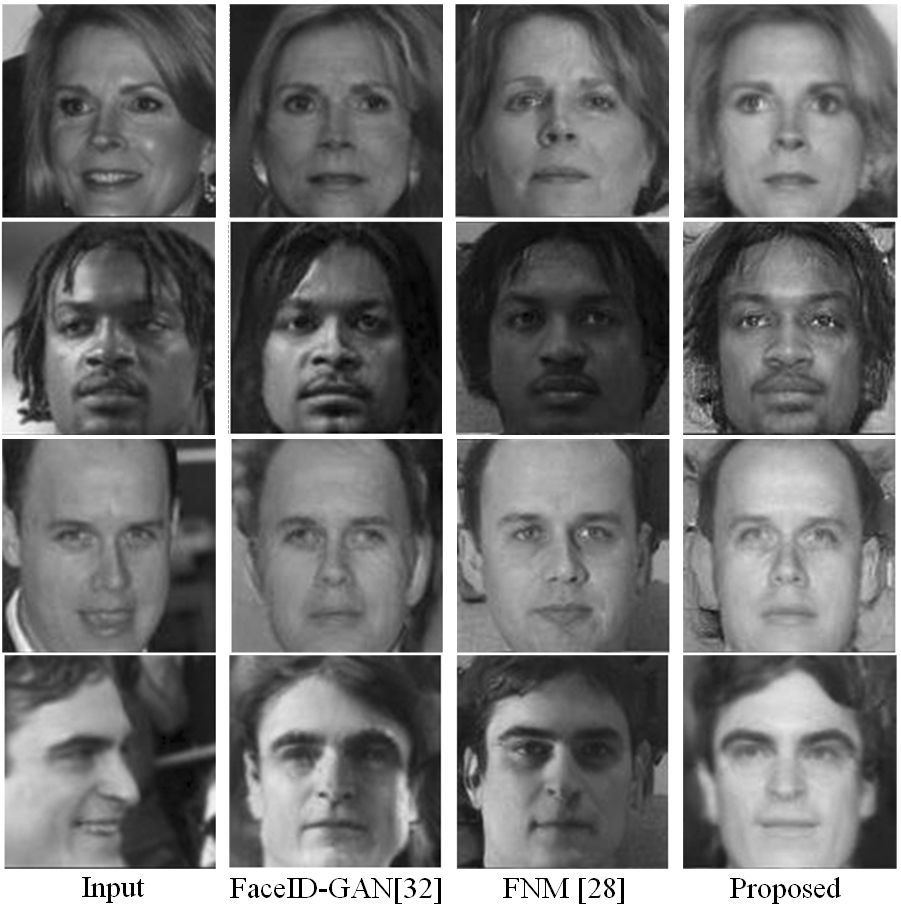} \par & \\ [-0.6cm]
\end{tabular}
\caption{
Face frontalization results derived by different methods.
}
\label{fig:LFW_vis}
\end{figure}

\begin{figure*}[!t]
\begin{tabular}{p{400pt}p{50pt}}
\centering
\includegraphics[width=0.95\textwidth,keepaspectratio=true]{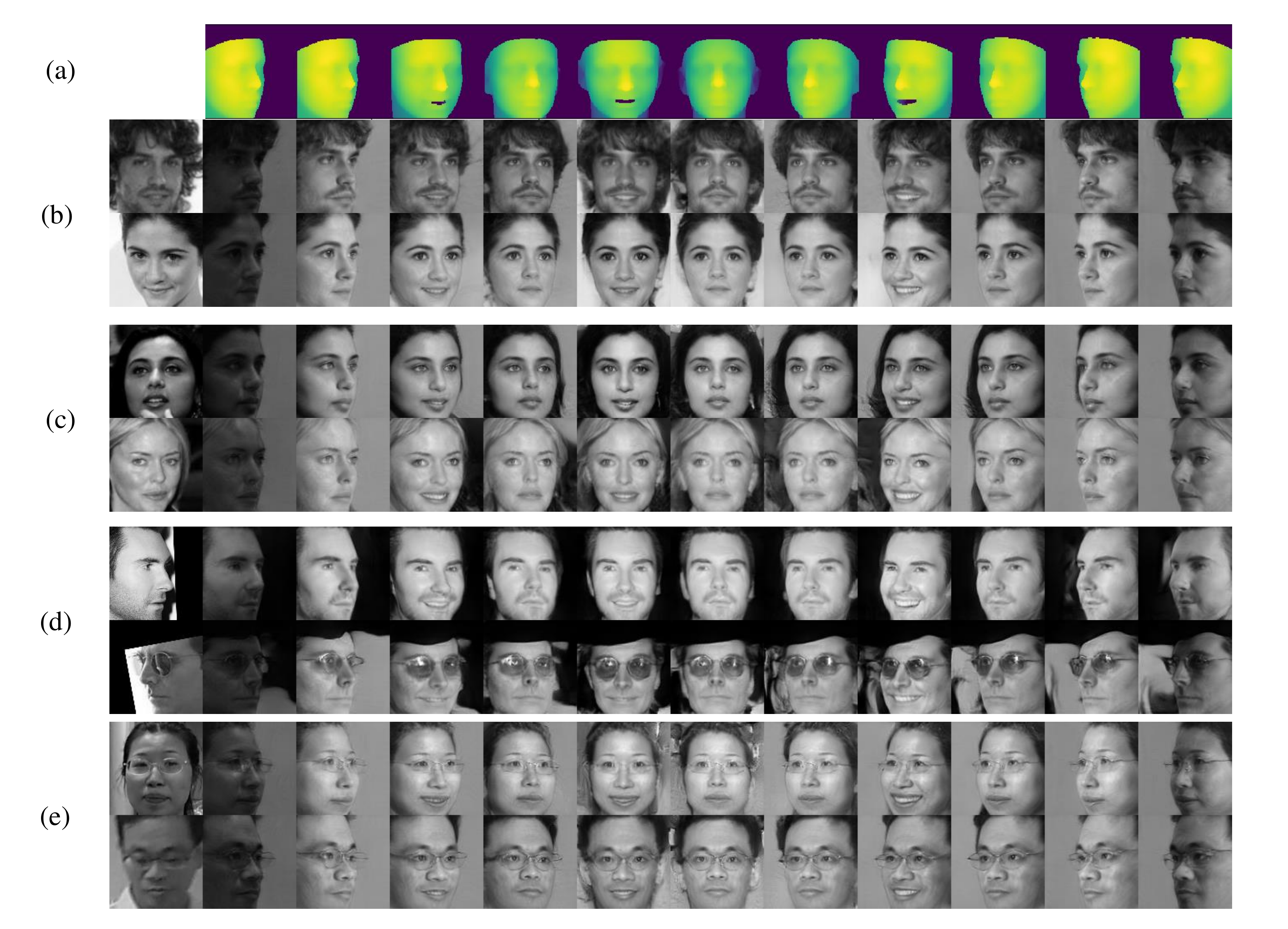} \par & \\ [-0.7cm]
\end{tabular}
\small
\caption{Synthesized faces for face samples  from  different datasets generated by DotFAN. 
The left-most column shows the inputs with random attributes (e.g., poses, expressions, and motion blurs). The top-most row illustrates 3D templates with specific poses and expressions. To guarantee the identity information of each synthetic face is observable, columns 3--11 show shadow-free results, and columns $2$ and $12$ show faces with shadows.
(a) 3D templates. (b) CelebA, (c) LFW, (d) CFP, and (e) SurveilFace.
}
\label{fig:rotation}
\end{figure*}

\subsection{Implementation Details}
\label{subsec420:implement}
Before training, we align the face images in the Multi-PIE and CASIA by MTCNN \cite{dai2016instance}. Structurally, our FEM is obtained by Resnet-50 pretained on MS-Celeb-1M~\cite{guo2016ms}, and FSR is implemented 
by a MobileNet~\cite{howard2017mobilenets} pretained on CASIA. 
To train DotFAN, each input face is resized to  $112 \times 112$. Both generator and discriminator exploit Adam optimizer \cite{kinga2015method} with $\beta_1=0.5$ and $\beta_2=0.999$. The total number of training iterations is $420,000$ with a batch-size of $28$, and the number of training epochs is $12$. 
The learning rate is initially set to be $10^{-4}$ and begins to decay after the $6$-th training epoch.

\subsection{Face Synthesis}
We verify the efficacy of DotFAN through the visual quality of i) face frontalization and ii) face rotation results.

\noindent \textbf{(A) Face frontalization:}
First, we verify if the identity information extracted from a frontalized face, produced by DotFAN, is of the same class as the identity of a given source face. Following \cite{shen2018faceid}, we measure the performance by using a face recognition model trained on MS-Celeb-1M. 
Next, we conduct frontalization experiments on LFW. 
Fig. \ref{fig:LFW_vis} illustrates the frontalization results derived by different methods; meanwhile, Tables \ref{tab:t1} and \ref{tab:t2} show the comparison of face verification results of frontalized faces.
%
This experiment set validates that i) 
compared with other methods, 
DotFAN achieves comparable visual quality in face frontalization, 
ii) shadows can be effectively removed by DotFAN, 
and iii) DotFAN outperforms the other methods in terms of verification accuracy, especially in the experiment on $\mbox{IJB-A}$ shown in Table \ref{tab:t2}, 
where DotFAN reports a much better TAR, i.e., $89.3\%$ on FAR@0.001 and $93.7\%$ on FAR@0.01, than existing approaches.

\begin{figure}[!t]
\centering
\begin{tabular}{p{200pt}p{30pt}}
\includegraphics[width=0.45\textwidth,keepaspectratio=true]{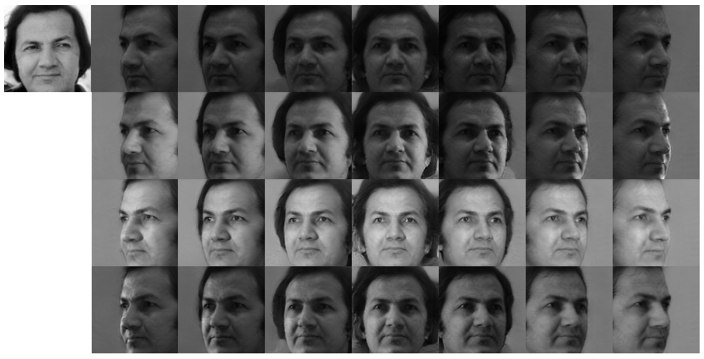}\par & \\ [-0.5cm]
\end{tabular}
\caption{Face augmentation examples (CelebA) containing augmented faces with $4$ illumination conditions and $7$ poses.
}

\label{fig:celeba_illumination}
\end{figure}

\noindent \textbf{(B) Face Rotation}
Fig. \ref{fig:rotation} demonstrates DotFAN's capability in synthesizing faces of given attributes, including pose angles, facial expressions, and shadows, 
while retaining the associated identities.
%
The source faces presented in the left-most column in Fig. \ref{fig:rotation} come from four datasets, i.e., CelebA, LFW, CFP~\cite{sengupta2016frontal}, and SurveilFace. 
CelebA and LFW are two widely-adopted face datasets; CFP contains images with extreme pose angles, e.g., $\pm90^\circ$; and, SurveilFace contains faces of variant illumination conditions and faces affected by motion-blurs. 
This experiment shows that DotFAN can stably synthesize visually-pleasing face images based on 3DMM parameters describing 3D templates. 
Finally, Fig. \ref{fig:celeba_illumination} shows some synthesized faces with shadows assigned with four different illumination codes. Note that all synthesized faces presented in this paper are produced by the same DotFAN model; no more data-oriented fine-tuning is required. 

\begin{table}[]
\begin{center}
\begin{tabular}{c c}
    \hline
    Method  & Verification Accuracy\\
    \hline
    HPEN~\cite{zhu2015high} & 96.25$\pm$0.76\\
    FF-GAN~\cite{yin2017towards} & 96.42$\pm$0.89\\
    FaceID-GAN~\cite{shen2018faceid} & 97.01$\pm$0.83   \\
    \hline
    Proposed & \textbf{99.18$\pm$0.39}\\
    \hline
\end{tabular}
\end{center}
	\caption{Verification accuracy on LFW.}
	\label{tab:t1}
	\vspace{-0.15in}
\end{table}

\begin{table}[]
\begin{center}
\begin{tabular}{c c c}
\hline
Method  & FAR@0.01 & FAR@0.001\\
\hline
PAM~\cite{masi2016pose}  & 73.3$\pm$1.8 & 55.2$\pm$3.2\\
DCNN~\cite{chen2016unconstrained} & 78.7$\pm$4.3 & -\\
DR-GAN~\cite{tran2017disentangled} & 77.4$\pm$2.7 & 53.9$\pm$4.3\\
FF-GAN~\cite{yin2017towards} & 85.2$\pm$1.0 & 66.3$\pm$3.3\\
FaceID-GAN~\cite{shen2018faceid} & 87.6$\pm$1.1 &69.2$\pm$2.7\\
\hline
Proposed & \textbf{93.7$\pm$0.5} & \textbf{89.3$\pm$1.0}  \\ 
\hline
\end{tabular}
\end{center}
	\caption{True-Accept-Rate (TAR) of verifications on IJB-A. 
	}
	\label{tab:t2}
	\vspace{-0.15in}
\end{table}
%
%
%
%
%
%

\subsection{Face Augmentation}
\vspace{-0.11in}

\begin{figure}[t]
\centering
\begin{tabular}{p{200pt}p{30pt}}
\includegraphics[width=0.47\textwidth,keepaspectratio=true]{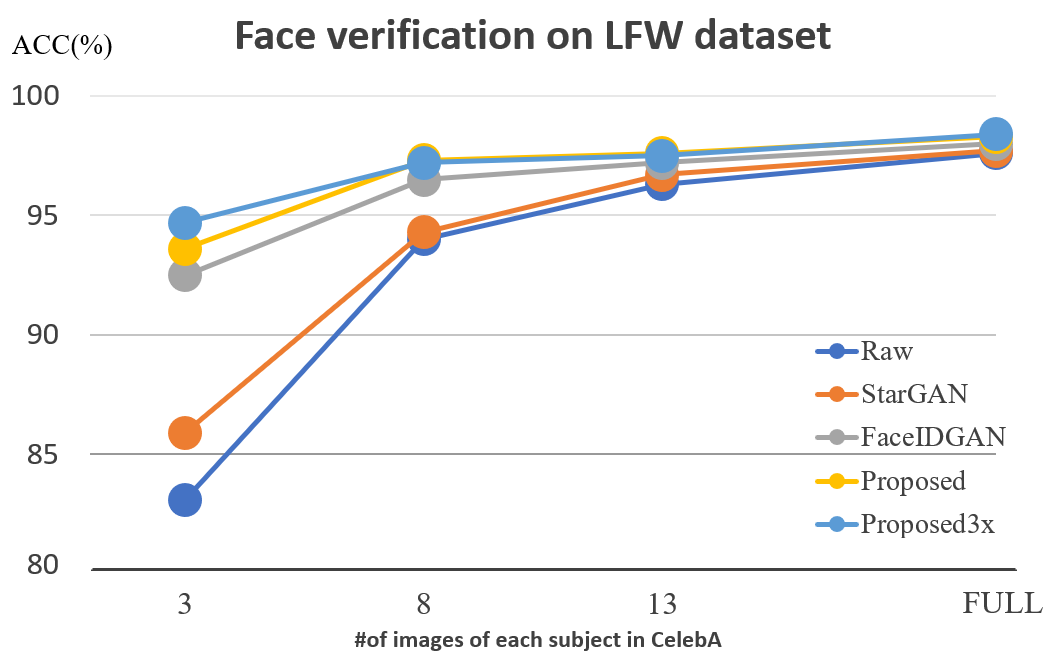} \par &\\ [-0.5cm]
\end{tabular}
\caption{Comparison of face verification accuracy on LFW  trained on different augmented dataset. The horizontal spacing highlights the size of raw training dataset sampled from CelebA.
}
\label{fig:trendofLFW}
\end{figure}

To evaluate the comprehensiveness of  domain-transferred augmentation by DotFAN, we perform data augmentation on the same dataset by using DotFAN, FaceID-GAN, and StarGAN first; then, we compare the recognition accuracy of different MobileFaceNet models \cite{chen2018mobilefacenets}, each trained on an augmented dataset, by testing them on LFW and SurveilFace.
%
%
%
StarGAN used in this experiment is trained on Mutli-PIE that is rich in illumination conditions; meanwhile, the FaceID-GAN is trained on CASIA to learn pose and expression representations.
%
%

\begin{table*}[!t]
    \vspace{-0.05in}
    \begin{center}
    \begin{scriptsize}
    \begin{tabular}{|c|c|c|c|c|c|c|c|c|} \hline
        Method & \multicolumn{2}{c|}{LFW} & \multicolumn{3}{c|}{SurveilFace-1} & \multicolumn{3}{c|}{SurveilFace-2} \\
        \cline{2-9} 
        & ACC & AUC & @FAR=0.001 & @FAR=0.01 & AUC & @FAR=0.001 & @FAR=0.01 & AUC\\ \hline
        \multicolumn{9}{|c|}{(a) \textbf{Sub-CelebA(3)} (totally $30,120$ images)} \\ \hline
        RAW       & 83.1 & 90.2 & 20.5 & 34.4 & 83.2 & 18.0 & 33.3 & 84.8  \\ \hline
        StarGAN   & 85.9 & 92.5 & 25.1 & 39.6 & 87.5 & 27.4 & 46.7 & 91.4  \\ \hline
        FaceID-GAN    & 
        \textcolor[rgb]{0,0.5,0}{92.5} & \textcolor[rgb]{0,0.5,0}{97.6} & \textcolor[rgb]{0,0.5,0}{34.6} & \textcolor[rgb]{0,0.5,0}{53.5} & \textcolor[rgb]{0,0.5,0}{92.8} & \textcolor[rgb]{0,0.5,0}{32.3} & \textcolor[rgb]{0,0.5,0}{54.0} & \textcolor[rgb]{0,0.5,0}{94.3}  \\ \hline
        Proposed 1x & \textcolor[rgb]{0,0,1}{93.6} & \textcolor[rgb]{0,0,1}{98.1} & \textcolor[rgb]{0,0,1}{35.7} & \textcolor[rgb]{0,0,1}{56.2} & \textcolor[rgb]{0,0,1}{93.6} & \textcolor[rgb]{0,0,1}{34.7} & \textcolor[rgb]{0,0,1}{57.8} & \textcolor[rgb]{0,0,1}{95.0} \\ \hline
        Proposed 3x  & \textcolor[rgb]{1,0,0}{94.7} & \textcolor[rgb]{1,0,0}{98.7} & \textcolor[rgb]{1,0,0}{36.8} & \textcolor[rgb]{1,0,0}{58.3} & \textcolor[rgb]{1,0,0}{94.6} & \textcolor[rgb]{1,0,0}{36.5} & \textcolor[rgb]{1,0,0}{60.8} & \textcolor[rgb]{1,0,0}{95.6} \\ \hline
        \multicolumn{9}{|c|}{(b) \textbf{Sub-CelebA(8)} (totally $75,796$ images)} \\ \hline 
        RAW       & 94.0 & 98.5 & 37.8 & 58.7 & 94.4 & 38.3 & 61.0 & 95.2  \\ \hline
        StarGAN   & 94.3 & 98.5 & 42.6 & 60.7 & 94.9 & 42.8 & 65.6 & 95.8  \\ \hline
        FaceID-GAN    &
        \textcolor[rgb]{0,0.5,0}{96.5} & \textcolor[rgb]{0,0.5,0}{99.3} & \textcolor[rgb]{0,0.5,0}{48.1} & \textcolor[rgb]{0,0.5,0}{65.6} & \textcolor[rgb]{0,0.5,0}{96.0} & \textcolor[rgb]{0,0.5,0}{45.7} & \textcolor[rgb]{0,0.5,0}{67.9} & \textcolor[rgb]{0,0.5,0}{96.8}  \\ \hline
        Proposed 1x & \textcolor[rgb]{1,0,0}{97.3} & \textcolor[rgb]{1,0,0}{99.5} & \textcolor[rgb]{1,0,0}{53.2} & \textcolor[rgb]{1,0,0}{71.2} & \textcolor[rgb]{1,0,0}{97.0} & \textcolor[rgb]{1,0,0}{49.1} & \textcolor[rgb]{1,0,0}{72.2} & \textcolor[rgb]{1,0,0}{97.2} \\ \hline
        Proposed 3x & 
        \textcolor[rgb]{0,0,1}{97.2} & \textcolor[rgb]{1,0,0}{99.5} & \textcolor[rgb]{1,0,0}{53.2} & 
        \textcolor[rgb]{0,0,1}{68.9} & 
        \textcolor[rgb]{0,0,1}{96.9} & 
        \textcolor[rgb]{0,0,1}{47.3} & 
        \textcolor[rgb]{0,0,1}{70.0} & 
        \textcolor[rgb]{0,0,1}{97.1} \\ \hline
        \multicolumn{9}{|c|}{(c) \textbf{Sub-CelebA(13)}  (totally $116,659$ images)} \\ \hline 
        RAW       & 96.3 & 99.1 & 47.4 & 67.8 & 96.2 & 43.5 & 67.0 & 96.5  \\ \hline
        StarGAN   & 96.7 & 99.3 & 48.3 & 68.1 & 96.7 & 46.3 & 70.0 & 96.7  \\ \hline
        FaceID-GAN    & \textcolor[rgb]{0,0.5,0}{97.2} & \textcolor[rgb]{0,0.5,0}{99.5} & \textcolor[rgb]{0,0.5,0}{53.3} & \textcolor[rgb]{0,0.5,0}{71.3} & \textcolor[rgb]{0,0.5,0}{97.0} & \textcolor[rgb]{0,0.5,0}{50.2} & \textcolor[rgb]{0,0,1}{72.3} & \textcolor[rgb]{0,0.5,0}{97.4}  \\ \hline
        Proposed 1x & \textcolor[rgb]{1,0,0}{97.6} & \textcolor[rgb]{0,0,1}{99.6} & \textcolor[rgb]{0,0,1}{56.2} & \textcolor[rgb]{0,0,1}{75.1} & \textcolor[rgb]{1,0,0}{97.7} & \textcolor[rgb]{0,0,1}{50.4} & \textcolor[rgb]{1,0,0}{73.9} & \textcolor[rgb]{0,0,1}{97.7}  \\ \hline
        Proposed 3x  & 
        \textcolor[rgb]{0,0,1}{97.5} & \textcolor[rgb]{1,0,0}{99.7} & \textcolor[rgb]{1,0,0}{56.7} & \textcolor[rgb]{1,0,0}{75.5} & \textcolor[rgb]{1,0,0}{97.7} & \textcolor[rgb]{1,0,0}{53.9} & \textcolor[rgb]{0,0.5,0}{72.2} & \textcolor[rgb]{1,0,0}{97.8} \\ \hline
        \multicolumn{9}{|c|}{(d) \textbf{CelebA (full CelebA dataset, $202,599$ images)}} \\ \hline 
        RAW       & 97.6  & 99.6 & 53.5 & 73.8 & 97.7 & 48.7 & 73.0 & 97.5  \\ \hline
        StarGAN   & 97.7 & 99.6 & 55.0 & 74.2 & 97.7 & 53.0 & 73.8 & 97.6 \\ \hline
        FaceID-GAN    & \textcolor[rgb]{0,0.5,0}{98.0} & \textcolor[rgb]{0,0,1}{99.7} & \textcolor[rgb]{0,0.5,0}{57.6} & \textcolor[rgb]{0,0.5,0}{76.4} & \textcolor[rgb]{0,0.5,0}{98.1} & \textcolor[rgb]{0,0.5,0}{54.1} & \textcolor[rgb]{0,0.5,0}{76.5} & \textcolor[rgb]{0,0,1}{98.0}  \\ \hline
        Proposed 1x & \textcolor[rgb]{0,0,1}{98.3} & \textcolor[rgb]{1,0,0}{99.8} & \textcolor[rgb]{1,0,0}{62.4} & \textcolor[rgb]{1,0,0}{80.9} & \textcolor[rgb]{1,0,0}{98.4} & \textcolor[rgb]{1,0,0}{57.1} & \textcolor[rgb]{0,0,1}{76.7} & \textcolor[rgb]{1,0,0}{98.1} \\ \hline
        Proposed 3x & \textcolor[rgb]{1,0,0}{98.4} & \textcolor[rgb]{0,0,1}{99.7} & \textcolor[rgb]{0,0,1}{61.4} & \textcolor[rgb]{0,0,1}{78.9} & \textcolor[rgb]{0,0,1}{98.2} & \textcolor[rgb]{0,0,1}{54.7} & \textcolor[rgb]{1,0,0}{77.8} &\textcolor[rgb]{0,0,1}{98.0} \\ \hline
    \end{tabular}
	\end{scriptsize}
	\end{center}
	\caption{
	Performance comparison among face recognition models trained on different datasets. Here, \textbf{Sub-CelebA($x$)} denotes a subset formed by randomly selecting $x$ images of each face subject from CelebA.
	}
	\label{tab:t3}
    \vspace{-0.2in}
\end{table*}

%
Table \ref{tab:t3} summarizes the results of this experiment set. We interpret the results focusing on Sub-experiment(a).
In Sub-experiment(a), we randomly select $3$ faces of each identity from CelebA to form the \textbf{RAW} training set, namely \textbf{Sub-CelebA(3)}, leading to about $30,000$ training samples in raw Sub-CelebA(3). 
The MobileFaceNet trained on raw Sub-CelebA(3) achieves a verification accuracy of $83.1\%$ on LFW, a true accept rate (TAR) of $20.5\%$ at FAR = $0.001$ on SurveilFace-1, and a TAR of $18.0\%$ at FAR = $0.001$ on SurveilFace-2. 
After generating about $30,000$ additional face images via DotFAN to double the size of training set, the verification accuracy on LFW becomes $93.6\%$, and the TAR values on SurveilFace datasets are all nearly doubled, as shown in the row named \textbf{Proposed 1x}. This experiment shows DotFAN is effective in face data augmentation and outperforms StarGAN and FaceID-GAN significantly. Furthermore, when we augment about $90,000$ additional faces to quadruple the size of training set, i.e., \textbf{Proposed 3x}, we have only a minor improvement in verification accuracy compared to \textbf{Proposed 1x}. This fact reflects that the marginal benefit a model can extract from the data diminishes as the number of samples increases when there is information overlap among data, as is what reported in \cite{cui2019class}. 

\begin{figure}[!t]
\centering
\begin{tabular}{p{200pt}p{40pt}}
\includegraphics[width=0.46\textwidth,keepaspectratio=true]{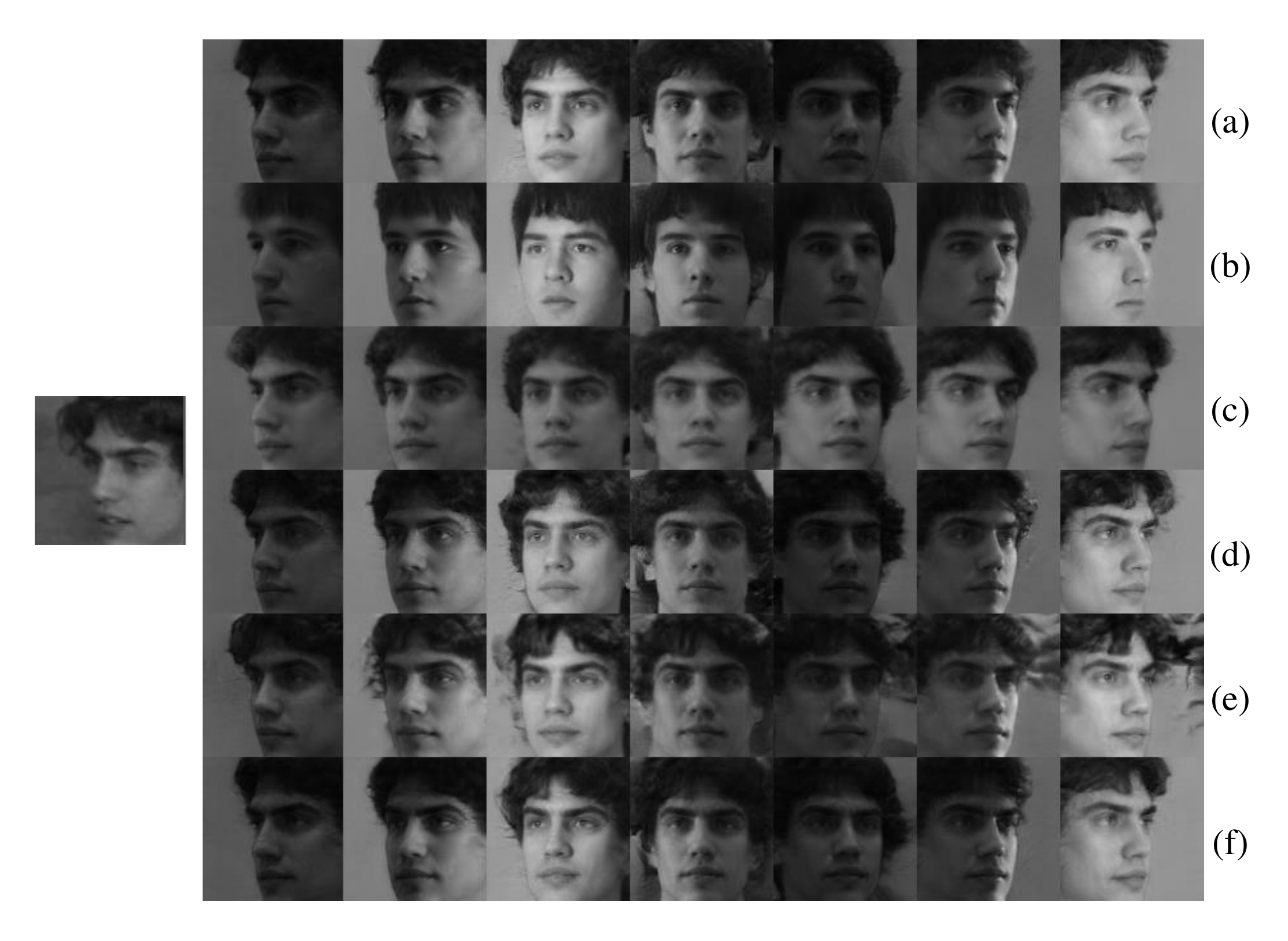} \par & \\ [-0.7cm]
\end{tabular}
\caption{Ablation study on loss terms. (a) Full loss. (b) w/o $L_{id}$, (c) w/o $L_{cls}$, (d) w/o $L_{cycle}$, (e) w/o $L_{pose}$, and (f) w/o $L_{sym}$. }
\label{fig:ablation_study}
\end{figure}

Consequently, Table \ref{tab:t3} and  Fig. \ref{fig:trendofLFW} reveals three remarkable points. 
First, although the improvement in verification accuracy decreases as the size of raw training set increases, DotFAN achieves significant performance gain on augmenting a small-size face training set, as demonstrated in all (RAW,~Proposed~1x) data pairs.
Second, the results obey \textit{the law of diminishing marginal utility} in Economics, as demonstrated in all (Proposed~1x,~Proposed~3x) data pairs. That is, a \textbf{1x} procedure is adequate to enrich a small dataset. Experiments also show that the \textbf{Proposed 3x} procedure seems to reach the upper-bound of data richness. 
Third, by integrating attribute controls on pose angle, illuminating condition, and facial expression with an identity-preserving design, DotFAN outperforms StarGAN and FaceID-GAN in domain-transferred face augmentation tasks. 

\subsection{Ablation Study}

In this section, we verify the effect brought by each loss term.
%
Fig. \ref{fig:ablation_study} depicts the faces generated by using different combinations of loss terms. The top-most row shows faces generated by using the full generator loss $\mathcal{L}_G$ described in Eq.(\ref{eq:totalloss}), whereas the remaining rows respectively show synthetic results derived without one certain loss term. 

As illustrated in Fig. \ref{fig:ablation_study}(b), without $\mathcal{L}_{id}$, DotFAN fails to preserve the identity information although other facial attributes can be successfully retained. By contrast, without $\mathcal{L}_{cls}$, DotFAN cannot control the illumination condition, and the resulting faces all share the same shadow (see Fig. \ref{fig:ablation_study}(c)). 
These two rows evidence that $\mathcal{L}_{cls}$ and $\mathcal{L}_{id}$ are indispensable in DotFAN design. 
Moreover, Fig. \ref{fig:ablation_study}(d) shows some unrealistic faces, e.g., a rectangular-shaped ear in the frontalized face; accordingly, $\mathcal{L}_{cycle}$ is important for photo-realistic synthesis. 
Finally, Fig. \ref{fig:ablation_study}(e)--(f) show that $\mathcal{L}_{pose}$ and $\mathcal{L}_{sym}$ are complementary to each other. As long as either of them functions, DotFAN can generate faces of different face angles. However, because $\mathcal{L}_{sym}$ is designed to learn only the mapping relationship between $+x^\circ$ face and $-x^\circ$ face by ignoring background outside the face region, artifacts may occur in the background region if $\mathcal{L}_{sym}$ works solely, as shown in  Fig. \ref{fig:ablation_study}(e).

%
%
%

\section{Conclusion}
\label{sec:conclud}


We proposed a Domain-transferred Face Augmentation network (DotFAN) for generating a series of variants of an input face image based on the knowledge of disentangled facial representation distilled from huge datasets. 
DotFAN is designed as a conditional CycleGAN with two extra subnetworks to learn the disentangled facial representation and produce a normalized face so that it can effectively generate face images of various facial attributes while preserving identity of synthetic images. 
Moreover, we proposed a pose-symmetric loss through which DotFAN can synthesize a pair of pose-symmetric face images directly at once. 
Extensive experiments demonstrate the effectiveness of DotFAN in augmenting small-size face datasets and improving their within-subject diversity. As a result, a better face recognition model can be learned from an enriched training set derived by DotFAN.



\ifCLASSOPTIONcaptionsoff
  \newpage
\fi



%

\bibliographystyle{IEEEtran}
\bibliography{facebib}

\appendices
\section*{Appendix}

\renewcommand{\thefigure}{S\arabic{figure}}
\renewcommand{\thetable}{S\arabic{table}}
\renewcommand{\thesection}{S\arabic{section}}

In this section, we show i) architectures of DotFAN's generator, general facial encoder, and discriminator, ii) face examples in the SurveilFace dataset, iii) DotFAN's capability for disentangled face representation, and iv) face images generated by DotFAN's data augmentation process.

\subsection{Model Architecture}
\label{sec:S01}
Listed in Table \ref{fig:E_A}, Table \ref{fig:G_A}, and Table \ref{fig:D_A} are the network structures of DotFAN's encoder ($E$), generator ($G$), and discriminator ($D$), respectively. 
Specified below are the notations used in Tables \ref{fig:E_A}-\ref{fig:D_A}. 
    \\$\bullet\;$\textbf{H}: Height of the input image.
    \\$\bullet\;$\textbf{W}: Width of the input image.
    \\$\bullet\;$\textbf{N}: Number of output channels.
    \\$\bullet\;$\textbf{K}: Kernel size.
    \\$\bullet\;$\textbf{S}: Stride size.
    \\$\bullet\;$\textbf{P}: Padding size.
    \\$\bullet\;$\textbf{Batch}: Batch normalization layer.
    \\$\bullet\;$\textbf{CONV}: 2D convolution layer.
    \\$\bullet\;$\textbf{FC}: Fully-connected layer.
    \\$\bullet\;$\textbf{TRANSPOSECONV}: 2D transpose convolution layer (for upsampling)
    \\$\bullet\;$\textbf{AvgPool}: Average pooling layer
    \\$\bullet\;$ Note that we set $\mathbf{H}=\mathbf{W}=112$ in all our experiments. 
%

\subsection{SurveilFace Dataset}
\label{sec:S02}
\begin{figure}[!t]
\centering
\includegraphics[width=0.48\textwidth,keepaspectratio=true]{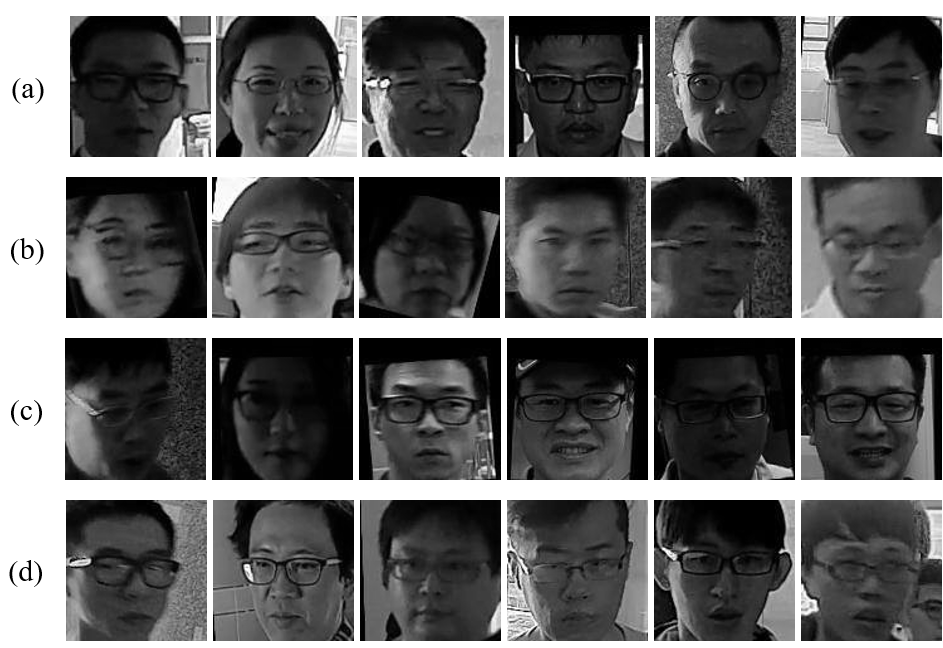} 
\caption{Image samples of SurveilFace dataset. Here we show four extreme conditions: (a) strong-backlight, (b) motion-blur, (c) extreme shadow, and (d) unconstrained viewpoint.}
\label{fig:Liteon_example}
\end{figure}

Demonstrated in Figure \ref{fig:Liteon_example} are face examples of the SurveilFace datasets. The two SurveilFace datasets were collected from a working-place surveillance system. Hence, uncontrolled real working environments may result in face photos affected by various extreme conditions, such as strong backlight, motion blurs, extreme shadows, or unconstrained viewpoints. These two datasets mimic the real-world conditions and thus are suitable for evaluating the face augmentation performance. 


\subsection{Disentangled Facial Representation}
\label{sec:S03}

\begin{figure*}
\centering
\includegraphics[width=1.0\textwidth,keepaspectratio=true]{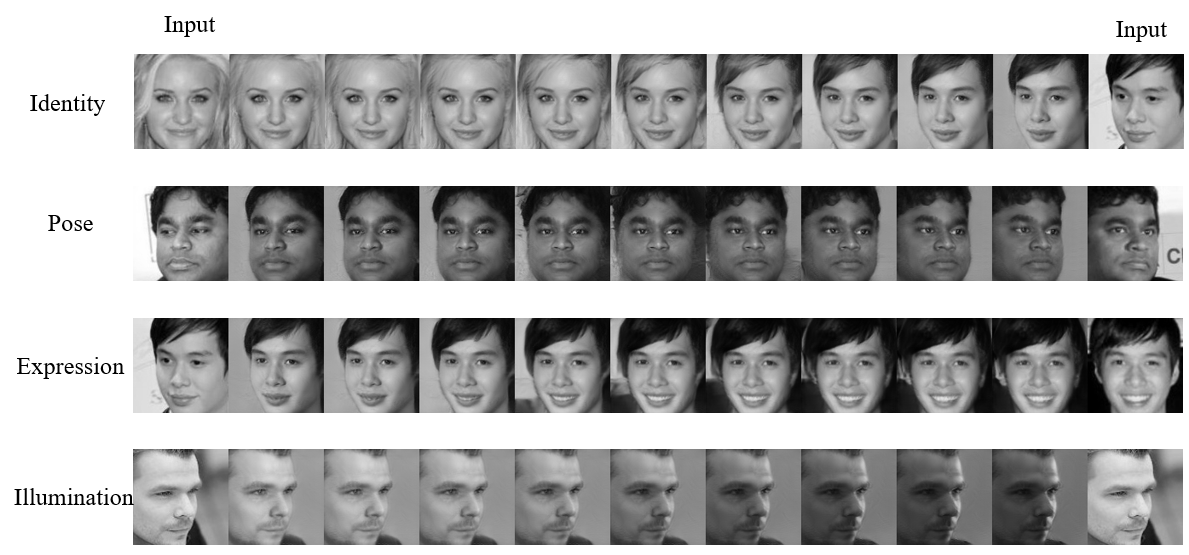} 
\caption{Disentangled facial representation. For each row, the morphing sequence is generated by using an attribute code $\mathbf{f}(\alpha)=\alpha \mathbf{f}_R+ (1-\alpha) \mathbf{f}_L$ with $0 \leq \alpha \leq 1$.
}
\label{fig:interpolation}
\end{figure*}

Figure \ref{fig:interpolation} exhibits synthesized faces to show DotFAN's capabilily for disentangled face representation. By exploiting our attribute code $\mathbf{f}=[l_\mathbf{x}, f_{id}, f_p, f_l]$, this experiment aims to show we can control the face synthesis result by manipulating $\mathbf{f}$.
In Figure \ref{fig:interpolation}, each row shows a sequence of faces. Each sequence was derived according to the convex combination---controlled by a scalar parameter $\alpha$---of two input attribute codes, i.e., $\mathbf{f}_R$ of the right-most face and $\mathbf{f}_L$ of the left-most. The sequence shown in the first row is derived by $\mathbf{f}_L = [l_{x}^l, f_{id}^l, f_{p}^l, f_l]$ and $\mathbf{f}_R = [l_{x}^r, f_{id}^r, f_{p}^r, f_l]$. With the illumination condition being fixed, we show that both the hairdo and the identity information vary smoothly with $\alpha$. 
The second row and the third row 
of Figure \ref{fig:interpolation} show the face interpolation results of controlled pose codes $f_{p}$. 
Because both pose information and expression information were encoded into $f_{p}$, these two sequences evidence that we can control the face synthesis by even editing only a segment of $\mathbf{f}$. 
Finally, 
the fourth row 
shows synthetic faces derived according to linearly interpolated illumination codes. Note that although the illumination code $f_l$ is a one-hot-vector, DotFAN can still approximate a shadow that varies almost linearly. 

\subsection{Data Augmentation}
\begin{figure*}
\centering
\includegraphics[width=1.0\textwidth,keepaspectratio=true]{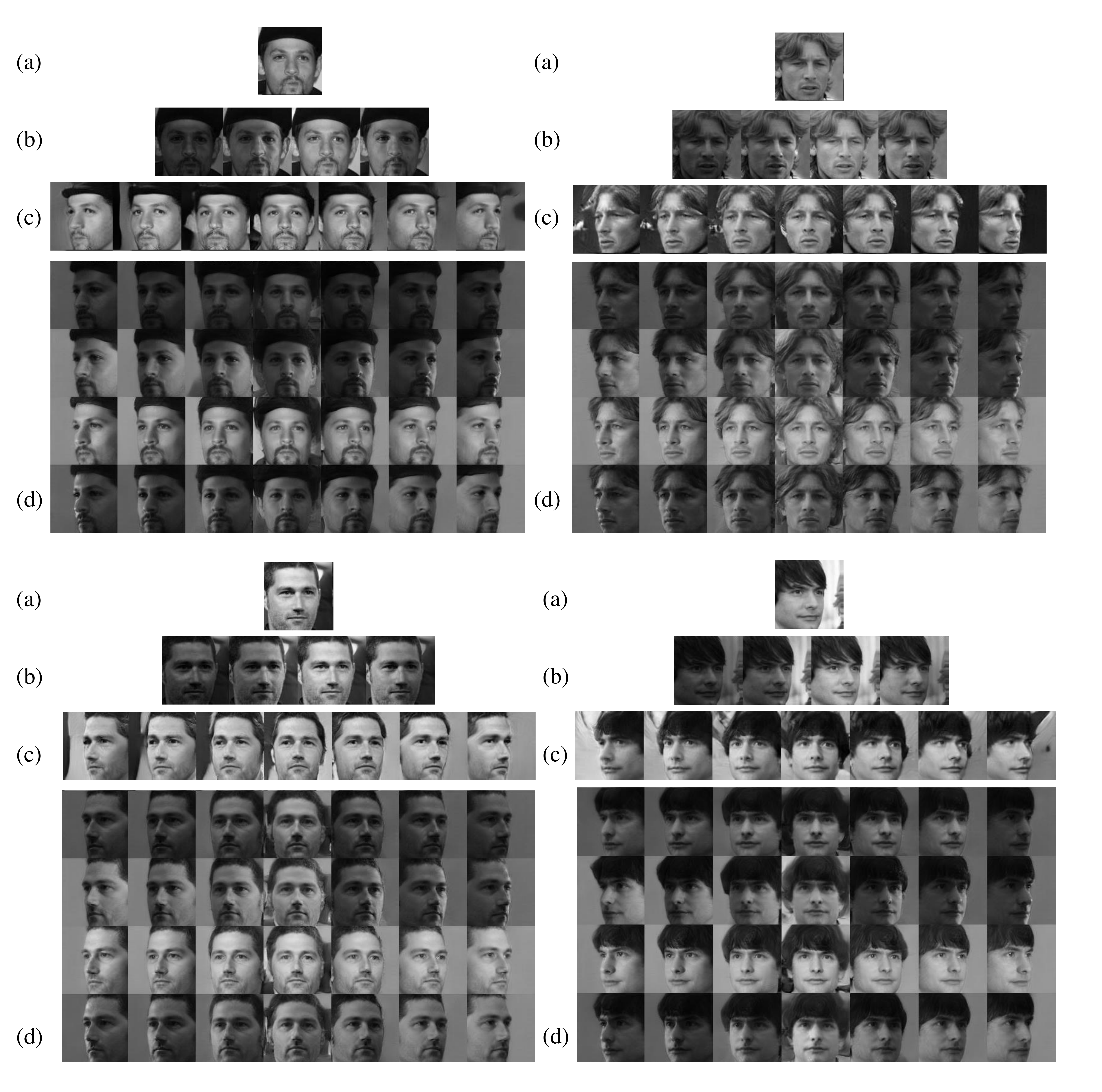} 
\caption{Augmentation examples of four face images from CelebA dataset. (a) input, (b) Stargan, (c) FaceID-gan, and (d) DotFAN}
\label{fig:app_example}
\end{figure*}

Finally, demonstrated in Figure \ref{fig:app_example} are face augmentation examples derived by different methods.

\begin{table*}
\vspace{-0.15in}
\begin{center}
	\includegraphics[width=0.8\linewidth]{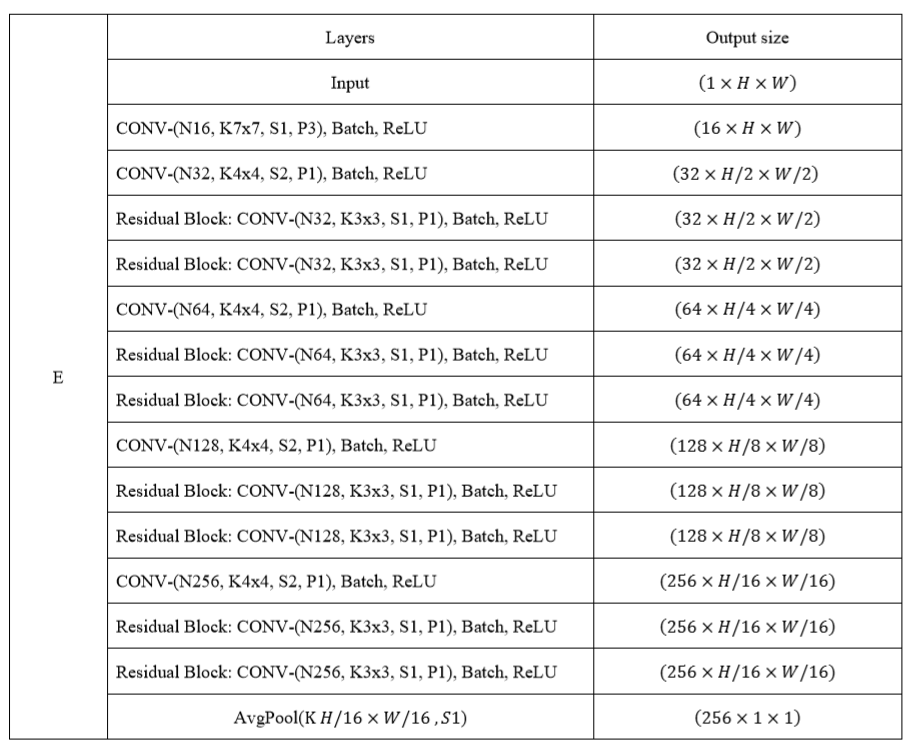}
\end{center}
\vspace{-0.15in}
\caption{Architecture of general facial encoder $E$.}	
\label{fig:E_A}
\end{table*}

\begin{table*}
\vspace{-0.15in}
\begin{center}
	\includegraphics[width=0.8\linewidth]{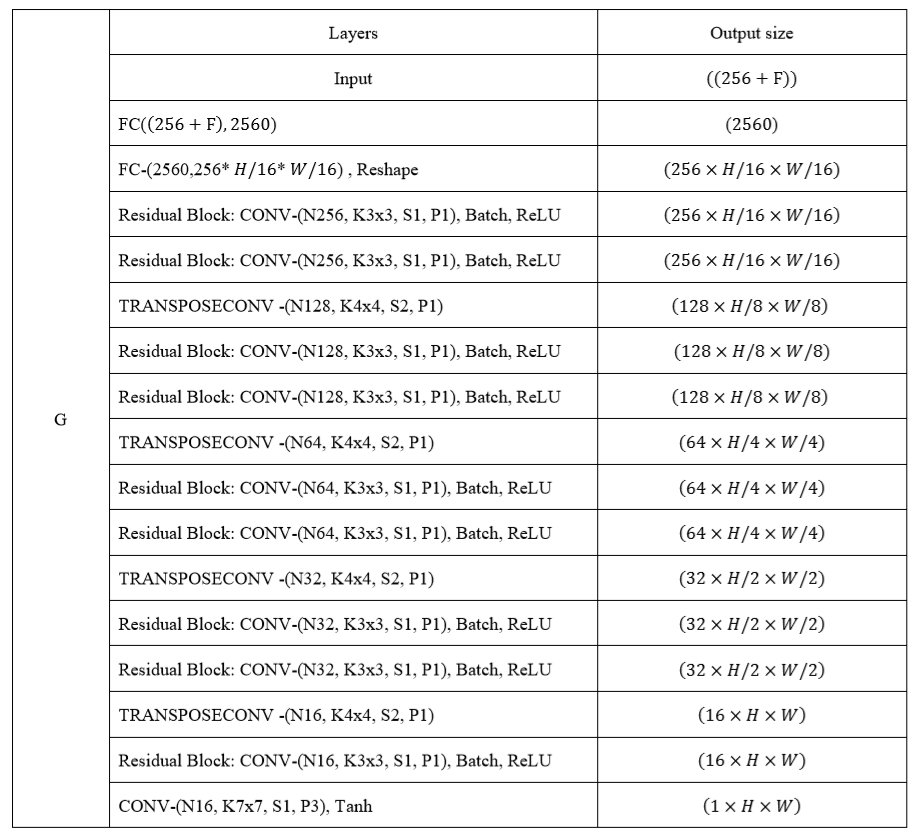}
\end{center}
\vspace{-0.15in}
\caption{Architecture of generator $G$.}	\label{fig:G_A}
\end{table*}

\begin{table*}
\vspace{-0.15in}
\begin{center}
	\includegraphics[width=0.8\linewidth]{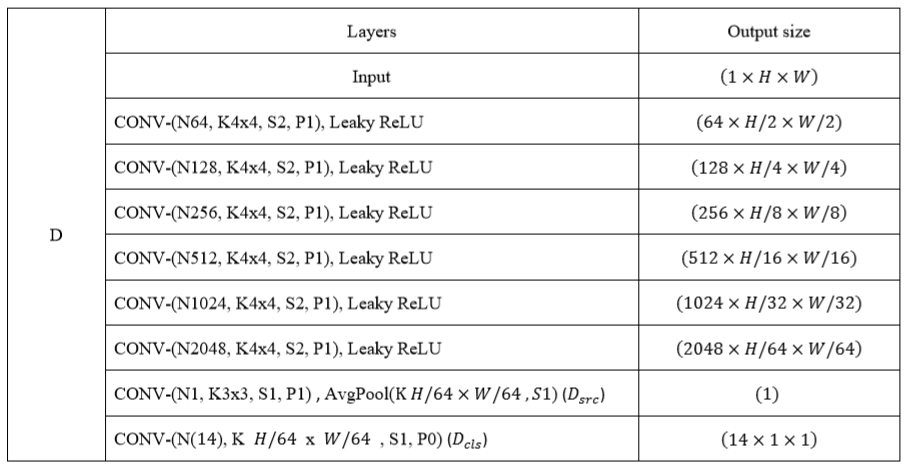}
\end{center}
\vspace{-0.15in}
\caption{Architecture of discriminator $D$.}	\label{fig:D_A}
\end{table*}

%
%

%








\end{document}